\documentclass{article}
\usepackage{amsmath,epsfig,bm, amssymb,subfigure,graphicx,algorithm,algorithmic,color}
\newtheorem{defi}{Definition}

\newtheorem{lemm}[defi]{Lemma}
\newtheorem{theo}[defi]{Theorem}

\newcommand{\proofend}{\hfill$\Box$\vspace{2mm}}

\newcommand{\argmin}{\mathop{\mathrm{argmin\,}}}

\newcommand{\Pnu}{P}
\newcommand{\Pde}{P'}

\newcommand{\mathbbE}{\mathbb{E}}

\newcommand{\mathbbR}{\mathbb{R}}

\newcommand{\boldzero}{{\boldsymbol{0}}}
\newcommand{\boldone}{{\boldsymbol{1}}}

\newcommand{\boldC}{{\boldsymbol{C}}}

\newcommand{\boldH}{{\boldsymbol{H}}}
\newcommand{\boldI}{{\boldsymbol{I}}}

\newcommand{\boldR}{{\boldsymbol{R}}}

\newcommand{\boldW}{{\boldsymbol{W}}}
\newcommand{\boldX}{{\boldsymbol{X}}}

\newcommand{\boldh}{{\boldsymbol{h}}}

\newcommand{\boldw}{{\boldsymbol{w}}}
\newcommand{\boldx}{{\boldsymbol{x}}}

\newcommand{\boldalpha}{{\boldsymbol{\alpha}}}

\newcommand{\boldtheta}{{\boldsymbol{\theta}}}

\newcommand{\boldmu}{{\boldsymbol{\mu}}}

\newcommand{\boldthetah}{{\widehat{\boldtheta}}}

\newcommand{\inputdim}{d}

\newcommand{\ratiosymbol}{r}
\newcommand{\ratio}{\ratiosymbol}

\newcommand{\ratiomodel}{\ratiosymbol_{\beta}}
\newcommand{\relratio}{\ratio_{\beta}}

\newcommand{\ptr}{p}
\newcommand{\pte}{p'}

\newcommand{\boldxtr}{\boldx}
\newcommand{\boldxte}{\boldx'}

\newcommand{\ntr}{n}
\newcommand{\nte}{n'}

\newcommand{\boldHh}{{\widehat{\boldH}}}
\newcommand{\boldhh}{{\widehat{\boldh}}}

\newcommand{\Hh}{{\widehat{H}}}
\newcommand{\hh}{{\widehat{h}}}

\usepackage{url,multirow}

\advance\oddsidemargin-1.0in
\date{\today}
\title{Interpreting Outliers: Localized Logistic Regression \\ for Density Ratio Estimation} %\thanks{Grants or other notes}

\author{
Makoto Yamada$^{1,2}$, Song Liu$^3$, Samuel Kaski$^4$\\
$^1$Kyoto University, Kyoto, Japan \\
$^2$PRESTO, JST \\
$^3$Institute of Statistical Mathematics, Tokyo, Japan \\
$^4$Aalto University, Espoo, Finland \\
\texttt{makoto.m.yamada@ieee.org}
}

\begin{document}
\maketitle

\begin{abstract}
We propose an inlier-based outlier detection method capable of both identifying the outliers and explaining why they are outliers, by identifying the outlier-specific features. Specifically, we employ an inlier-based outlier detection criterion, which uses the ratio of inlier and test probability densities as a measure of plausibility of being an outlier. For estimating the density ratio function, we propose a localized logistic regression algorithm. Thanks to the locality of the model, variable selection can be outlier-specific, and will help interpret why points are outliers in a high-dimensional space. Through synthetic experiments, we show that the proposed algorithm can successfully detect the important features for outliers. Moreover, we show that the proposed algorithm tends to outperform existing algorithms in benchmark datasets.
\end{abstract}

\section{Introduction}
Outlier detection (a.k.a Anomaly detection), the problem of finding anomalies in data, is attracting a lot of attention in the machine learning and data mining communities \cite{aggarwal2001outlier}. It is needed in various real-world applications including spacecraft anomaly detection, computer fault detection, and intrusion detection in network systems \cite{fujimaki2005approach,ide2004eigenspace,yamanishi2000line}. Recently, the problem of outlier detection from high-dimensional data such as multi-sensor data has been attracting increasing attention. Moreover, the demand for \emph{interpretable} models of data is increasing, and it would be highly useful if models used to detect outliers would be interpretable as well; interpretability could be achieved by models that select descriptive features. However, to the best of our knowledge, there is no existing framework to select features of outliers. 

%Unsupservised outlier detection (One-class SVM, LOF, etc.)
Unsupervised outlier detection is a common approach to detect anomalies. For example, one-class support vector machines (OSVM) \cite{book:Schoelkopf+Smola:2002}, kernel density estimation (KDE) \cite{sheather1991reliable}, and  local outlier factors (LOF) \cite{breunig2000lof} are widely used unsupervised outlier detection algorithms. The key idea of these approaches is to regard samples which are located in low-density region as outliers. Unsupervised algorithms perform well if the low-density assumption holds. If the low-density assumption does not hold, new approaches are needed, and in any case models need to be made interpretable which is not straightforward in kernel methods in particular.

%inlier-based outlier detection (Logistic regression, KLIEP, uLSIF/RuLSIF).
If the low-density assumption does not hold, or does not result in high enough detection accuracy, 
inlier-based outlier detection is useful \cite{hido2008inlier,smola2009relative,KAIS:Hido+etal:2011,chakhchoukh2016statistical}. The inlier-based methods model known normal samples (inliers), and detect deviations in test samples. 
Since additional knowledge is available instead of just the test samples, in the form of some training samples being known to be normal, inlier-based methods tend to have higher detection accuracies than unsupervised algorithms.  A widely used inlier-based method is based on the density-ratio between inlier and test densities, and uses the density-ratio as a measure of plausibility of data being outliers. To estimate the density-ratio, a number of \emph{direct} density-ratio estimation algorithms have been proposed, based on logistic regression \cite{bickel2007discriminative}, a Kullback-Leibler importance estimation procedure (KLIEP) \cite{sugiyama2008direct}, and (relative) unconstrained least-squares importance fitting (uLSIF/RuLSIF) \cite{JMLR:Kanamori+etal:2009,yamada2011relative}. The key idea of the direct estimation methods is to directly estimate the density ratio function without estimating probability densities. The inlier-based outlier detection algorithms empirically outperform unsupervised counterparts. However, since the direct methods employ kernel models, it is hard to interpret the detected outliers.

In this paper, we propose an inlier-based outlier detection method, which incorporates feature selection into the outlier detection algorithm. More specifically, we model each sample $\boldx_i \in \mathbbR^d$ by introducing a locally linear density-ratio model $\exp(\boldw_i^\top \boldx_i)$, where $\boldw_i \in \mathbbR^d$ is the model parameter vector for the sample $\boldx_i$. Since the model is locally linear, we can apply well-functioning feature selection methods to make the local models interpretable in terms of a small set of features, by properly regularizing the  model parameter vector $\boldw_i$. To estimate the locally linear density ratio function, we propose a convex optimization approach. In summary, the proposed approach has two compelling properties: the model is interpretable in terms of features automatically selected, and convex optimization guarantees a globally optimum solution.  We first illustrate the proposed method on a synthetic data set, and then compare it with state-of-the-art outlier detection methods on benchmark data.

\noindent {\bf Contribution:} 
\begin{itemize}
\item We propose a new locally linear density-ratio based outlier detection method.
\item We propose a convex optimization algorithm, which guarantees a globally optimal solution.
\item The outliers can be interpreted in terms of a small number of features, which is possible due to employing the locally linear model. This is highly useful for practitioners.
\end{itemize}

\section{Proposed Method}
In this section, we first formulate the outlier detection problem as density-ratio estimation and then introduce the new method for inlier-based outlier detection based on locally linear density-ratio estimation.
\subsection{Problem Formulation}
Let us denote an input vector by $\boldx = [x^{(1)}, \dots, x^{(d)}]^\top \in \mathbbR^d$. The set of samples $\{\boldxtr_i\}_{i = 1}^{\ntr}$ has been drawn i.i.d. from a probability density $p(\boldx)$ and another set of samples is drawn i.i.d. from an another probability density $p'(\boldx)$. In  inlier-based outlier detection, we assume that $p'(\boldx)$ is the probability density for inliers and $p(\boldx)$ is the probability density for test data (assumed to be a mixture of inliers and outliers).

The goal of this paper is to estimate a density-ratio function from the observed samples as
\begin{align}
\label{eq:ratio}
r(\boldx) = \frac{p'(\boldx)}{p(\boldx)}.
\end{align}
 %In particular, we aim to learn a model with an interpretable sparsity pattern in the features.

In particular, we aim to use the ratio function to detect outliers. Specifically, we detect an outlier as
\begin{eqnarray}
\left\{
\begin{array}{cc}
r(\boldx) > \tau & (\textnormal{Inlier}) \\
r(\boldx) \leq \tau & (\textnormal{Outlier})
\end{array}
\right..
\end{eqnarray}
where $\tau \geq 0$ is a thresholding parameter. 
Moreover, we focus on selecting a set of features for the outliers.  %To the best of our knowledge, this is the first work to select features of outliers. 

\subsection{Locally Linear Density-Ratio}
The goal of this paper is to not only detect outliers but also to select features of the outliers. To this end, we employ a locally linear model to estimate the density-ratio function Eq.\eqref{eq:ratio}.

Let us define the probability densities as
\begin{align*}
\ptr(\boldx) &= p(\boldx | y = -1), \\
\pte(\boldx) &= p(\boldx | y = +1). 
\end{align*}
Then the ratio function $r(\boldx)$ can be written as
\begin{align*}
r(\boldx) &= \frac{\pte(\boldx)}{\ptr(\boldx)} = \frac{p(y= -1)}{p(y = +1)} \frac{p(y = +1|\boldx)}{p(y = -1|\boldx)}. 
%\frac{\frac{\pte(\boldx)}{\pte(\boldx) + \ptr(\boldx)}}{\frac{\ptr(\boldx)}{\pte(\boldx) + \ptr(\boldx)}}.
\end{align*}
%and take there ratio.

To obtain $p(y = +1|\boldx)$ and $p(y = -1|\boldx)$, we can use any probabilistic classifier. In this paper, we employ a logistic regression model.  More specifically, we train logistic regression by setting $\{\boldxtr_i\}_{i = 1}^{\ntr}$ as the samples of the  positive class and $\{\boldxte_j\}_{j = 1}^{\nte}$ as the samples of  the negative class. Since logistic regression is a linear method, it would obviously  perform poorly on nonlinear data, and kernel logistic regression (KLR) \cite{zhu2005kernel} could be used instead. However, interpretability is a known weakness of kernel methods.

To make the method interpretable by selecting features for samples, in addition to having a high detection power stemming from nonlinearity, we introduce a locally  linear density-ratio model:
\begin{align}
\label{eq:linmodel}
\begin{split}
p(y = +1 | \boldx_i; \boldw_i)  &= \frac{1}{1 + \exp(-\boldw_i^\top \boldx_i)}, \\
p(y = -1 | \boldx_i; \boldw_i)  &= \frac{\exp(-\boldw_i^\top \boldx_i)}{1 + \exp(-\boldw_i^\top \boldx_i)},
\end{split}
\end{align}
where $\boldw_i \in \mathbbR^{d}$ contains the regression coefficients for the sample $\boldx_i$ and $^\top$ denotes the  transpose. Note that this model is a local variant of logistic regression. If the weight vectors were equal, $\boldw = \boldw_1 = \ldots = \boldw_n$, the locally linear density-ratio would reduce to a simple logistic regression model.

The ratio function for $\boldx_i$ is given as
\begin{align}
\label{eq:outlier_score}
r(\boldx_i; \boldw_i) %&= \frac{p(y= -1)}{p(y = +1)} \frac{p(y = +1|\boldx)}{p(y = -1|\boldx)} \\
&=\frac{\ntr}{\nte}  \frac{p(y = +1|\boldx_i; \boldw_i)}{p(y = -1|\boldx_i; \boldw_i)} =\frac{\ntr}{\nte} \exp(\boldw_i^\top \boldx_i).
\end{align}
 
The key challenge of this \emph{local} model is that there are as many unknown variables as observed variables in Eq. \eqref{eq:linmodel}. Thus, we need to regularize the model parameters to obtain a good solution.

\subsection{Localized Logistic Regression}
Let us denote the pooled paired data $\{(\widetilde{\boldx}_i, y_i)\}_{i = 1}^{\ntr + \nte}$, where $y_i = 1$ if $\widetilde{\boldx}_i$ is a sample of $\{\boldx_i\}_{i = 1}^{\ntr}$ and $y_i = -1$ otherwise. We estimate the model parameters $\boldw_i$ via solving the following optimization problem:
\begin{align}
\label{eq:llr}
\min_{\boldW} \hspace{0.1cm} J(\boldW) = \sum_{i = 1}^{\ntr + \nte} \log (1 + \exp(-y_i\boldw_i^\top \widetilde{\boldxtr}_i)) + \rho(\boldW; \boldR,\lambda_1,\lambda_2),
\end{align}
where 
\begin{align*}
%\label{eq:struct_reg}
\rho(\boldW; \boldR,\lambda_1,\lambda_2) \!=\! \lambda_1 \!\sum_{i,j = 1}^{n} r_{ij} \|\boldw_i - \boldw_j\|_{2} \!+\! \lambda_2 \!\sum_{i = 1}^{\ntr}\|\boldw_i\|_1^2,
%\lambda_1 \|\boldW\|_{2,1} + \lambda_2 \|\boldW^\top\|_{1,2}.
\end{align*}
is an exclusive fused regularization term \cite{yamada2017aistats}, which is a mixture of the network regularizer \cite{hallac2015snapvx} and the exclusive regularizer \cite{kowalski2009sparse,zhou2010exclusive,kong2014exclusive}.

Here $\lambda_1 \geq 0$ and $\lambda_2 \geq 0$ are the regularization parameters.  By imposing the network regularization, we regularize the model parameters $\boldw_i$ and $\boldw_j$ to be similar if $r_{ij} > 0$. In this paper, we use the following similarity function:
\[r_{ij} =\left\{ \begin{array}{ll}
\delta_{ij} \exp \left(-\frac{\|\widetilde{\boldx}_i - \widetilde{\boldx}_j\|^2}{2\sigma^2}\right) & \text{if}~\widetilde{\boldx}_j~\text{is the neighbor of} ~\widetilde{\boldx}_i \\
0 & \textnormal{otherwise,}
\end{array} \right. 
\]
where $\sigma^2$ is the kernel parameter. Here $\delta_{ij} = 1$ if $\boldx_j$ is included in the $K$ closest neighbors of $\widetilde{\boldx}_i$, otherwise $\delta_{ij} = 0$. Eq. \eqref{eq:llr} is convex and hence has a globally optimal solution.

If $\lambda_1$ is large, we can cluster the samples according to how similar the  $\boldw_i$s are.  More specifically,  the $i$-th sample and $j$-th sample belong to the same cluster if $\|\boldw_i - \boldw_j\|_2$ is small (possibly zero).

The second regularization term in $\rho(\boldW; \boldR, \lambda_1, \lambda_2)$ is the $\ell_{1,2}$ regularizer (a.k.a., exclusive regularizer) \cite{kowalski2009sparse,zhou2010exclusive,kong2014exclusive}. By imposing the $\ell_{1,2}$ regularizer, we can make the model select a small number of elements within each $\boldw_i$ to be non-zero. Since we have the $\ell_2$ norm over the weight vectors, the $\boldw_i$ remain non-zero (i.e., $\boldw_i \neq \boldzero$).  The $\boldw_i$ can be easily interpretable by checking similarities and differences in the sparsity patterns, more easily than in dense vectors. %Note that while simply imposing the $\ell_1$ regularizer for all weights would induce sparsity too, for the heavy regularization required due to the small sample size, many of the $\boldw_i$ would be shrunk to zero. See Figure \ref{fig:toy} for an example.

%(first term in the following equation): %\ref{eq:struct_reg}):

\begin{algorithm}[t]
\caption{Localized Logistic Regression for outlier detection (LLR)}
\label{alg:alg}
\begin{algorithmic}
\STATE Input: $\boldX \in \mathbbR^{d \times n}$, $\boldX'\in \mathbbR^{d \times n'}$, $\boldR \in \mathbbR^{n \times n}$, $\lambda_1$, and $\lambda_2$.
\STATE Output: outlier score $r(\boldx'_i;\widehat{\boldw}_i)$.
\STATE Set $t = 0$, Initialize $\boldC_g^{(0)}$, $\boldC_e^{(0)}$.%, $\boldU = [~]$
\REPEAT
\STATE  ${\boldW}^{(t+1)} = \argmin_{\boldW}\widetilde{J}(\boldW)$
\STATE Update $\boldC_g^{(t+1)}$ and $\boldC_e^{(t+1)}$.
\STATE $t = t + 1$.
\UNTIL{Converges}
\RETURN $r(\boldx'_i; \widehat{\boldw}_i) = \exp(\widehat{\boldw}_i^\top \boldx'_i), i = 1, \ldots, n'$.
\end{algorithmic}
\end{algorithm}

\subsection{Optimization}
To solve Eq. \eqref{eq:llr}, we employ an iterative re-weighted updating technique. 
 Instead of optimizing Eq.~\eqref{eq:llr}, we equivalently optimize
\begin{align}
\label{eq:objective_function3}
\widetilde{J}(\boldW) &= \sum_{i = 1}^{\ntr + \nte} \log (1 + \exp(-y_i\boldw_i^\top \widetilde{\boldxtr}_i))  + \lambda_1  \text{tr}(\boldW \boldC^{(t)}_g\boldW^\top) \nonumber \\ 
&\phantom{=}+ \lambda_2\boldone_d^\top (\boldW \circ (\boldC^{(t)}_e \circ \boldW))\boldone_n,
\end{align}
where $\boldC_g^{(t)} \in \mathbbR^{(\ntr + \nte) \times (\ntr + \nte)}$ and $\boldC_e^{(t)} \in \mathbbR^{d \times (\ntr + \nte)}$ are defined as\footnote{When $\boldw_i - \boldw_j = \boldzero$, then $\boldC_g$ is the subgradient of $\sum_{i,j=1}^{\ntr + \nte} r_{ij}\|\boldw_i - \boldw_j\|_2$. Also, $\boldC_e$ is the subgradient of $\sum_{i = 1}^{\ntr + \nte} \|\boldw_i\|_{1}^2$ when $[|\boldW|\|_{i,j} = 0$. However, we cannot set elements of $\boldC_g$ to 0 (i.e.,  when $\boldw_i - \boldw_j = \boldzero$) or the element of $[\boldC_e]_{i,j} = 0$ (i.e., when $|\boldW_{ij}| = 0$), otherwise the Algorithm~1 cannot be guaranteed to converge. To deal with this issue, we can use $\sum_{i,j=1}^{\ntr + \nte} r_{ij}\|\boldw_i - \boldw_j + \epsilon \|_2$ and $\sum_{i = 1}^{\ntr + \nte} \|\boldw_i + \epsilon\|_{1}^2$ ($\epsilon > 0$) instead \cite{kong2014exclusive,nie2010efficient}. }
\begin{align*}
[\boldC_g^{(t)}]_{i,j} &\!=\!\left\{\!\! \begin{array}{ll}
\sum_{j' = 1}^{\ntr + \nte} \!\frac{r_{ij'}}{\|\boldw_{i}^{(t)} - \boldw_{j'}^{(t)}\|_{2}} \!-\! \frac{r_{ij}}{\|\boldw_i^{(t)} \!-\! \boldw_j^{(t)}\|_{2}} & (i = j) \\
\frac{-r_{ij}}{\|\boldw_i^{(t)} - \boldw_j^{(t)}\|_{2}} & (i \neq j) 
\end{array} \right. \\
[\boldC_e^{(t)}]_{i,j} &\!=\! \frac{\|\boldw_{i}^{(t)}\|_1}{|[\boldW^{(t)}]_{i,j}|}.
\end{align*}
We can easily show that this objective function is convex, which guarantees a globally optimal solution.

We propose to use an iterative approach to optimize Eq.~\eqref{eq:objective_function3}. With given  $\boldC_g^{(t)}$ and $\boldC_e^{(t)}$, the optimal solution of $\boldW$ is obtained by solving $\min_{\boldW} \widetilde{J}(\boldW)$, and the obtained solution satisfies $\widetilde{J}(\boldW^{(t+1)}) \leq \widetilde{J}(\boldW^{(t)})$ since $\widetilde{J}(\boldW)$ is a convex function. For optimization We use conjugate gradients.

\subsection{Convergence Analysis} 
Next, we prove the convergence of the algorithm. This theorem is analogous to Theorem 1 in \cite{yamada2017aistats}. 

\begin{lemm}
Under the updating rule of Eq.\eqref{eq:objective_function3},
\[
\widetilde{J}(\boldW^{(t+1)}) - \widetilde{J}(\boldW^{(t)}) \leq 0.
\]
\noindent Proof: Since the objective function of Eq.\eqref{eq:objective_function3} is a convex function and the optimal solution is obtained as ${\boldW}^{(t+1)} = \argmin_{\boldW} \widetilde{J}(\boldW)$, the obtained solution $\boldW^{(t+1)}$ is the global solution. That is $\widetilde{J}(\boldW^{(t+1)}) - \widetilde{J}(\boldW^{(t)}) \leq 0$.
\end{lemm}

\begin{lemm} \cite{yamada2017aistats}
\[
{J}(\boldW^{(t+1)}) - {J}(\boldW^{(t)})  \leq \widetilde{J}(\boldW^{(t+1)}) - \widetilde{J}(\boldW^{(t)}).
\]
%\noindent Proof: See \cite{yamada2017aistats}
\end{lemm}

\begin{theo}
\label{theo:theo1}
The Algorithm 1 will monotonically decrease the objective function Eq.~\eqref{eq:llr} in each iteration.

\vspace{.1in}
\noindent Proof: Under the updating rule of $\widehat{\boldW}^{(t+1)} = \argmin_{\boldW} \widetilde{J}(\boldW)$, we have the following inequality using Lemma 1 and Lemma 2:
\begin{align*}
{J}(\boldW^{(t+1)}) - {J}(\boldW^{(t)}) \leq \widetilde{J}(\boldW^{(t+1)}) - \widetilde{J}(\boldW^{(t)}) \leq 0.
%J(\boldW^{(t+1)}) \leq J(\boldW^{(t)}).
\end{align*}
That is, the Algorithm 1 will monotonically decrease the objective function of Eq.~\eqref{eq:llr}.  \proofend 
\end{theo}

\begin{table*}[t!]
\centering
\caption{Comparison of outlier detection algorithms. 
}
\small
\label{tb:method_comp}
\begin{tabular}{|c|c|c|c|c|c|c|}
\hline
      & OSVM & LOF & $\ell_1$-LR & KLIEP & uLSIF & Proposed \\ \hline
Outlier detection model & Unsupervised & Unsupervised & {\bf Inlier-based} & {\bf Inlier-based} & {\bf Inlier-based} & {\bf Inlier-based} \\ \hline
Linear/Nonlinear & {\bf Nonlinear} & {\bf Nonlinear} & Linear & {\bf Nonlinear} & {\bf Nonlinear} & {\bf Nonlinear} \\ \hline
Feature Selection & No & No & {\bf Yes} & No  & No & {\bf Yes} \\ \hline
Ratio-Based & No & No & {\bf Yes} & {\bf Yes} & {\bf Yes} & {\bf Yes} \\ \hline
\end{tabular}
\vspace{-.1in}
\end{table*}

\section{Related Work}
Our work is  related to the fields of \emph{unsupervised outlier detection} and \emph{inlier-based outlier detection}. Table~\ref{tb:method_comp} summarizes the comparison of outlier detection algorithms. Below, we review some important related methods in each subarea.

\subsection{Unsupervised Outlier Detection}
Suppose we have a set of samples $\{\boldx_i\}_{i =1}^n$. The goal of unsupervised outlier detection methods is to find outliers without knowing more of the $\{\boldx_i\}_{i =1}^n$.

\vspace{.1in}
\noindent {\bf Kernel Density Estimation (KDE)} \cite{latecki2007outlier}
A popular outlier detection method is a probability density based approach. The key idea is to regard a sample which has low probability in the training set as an anomalous sample. 

KDE is a widely used non-parametric technique to estimate a density $p(\boldx)$ from samples $\{\boldx_k\}_{k = 1}^n$. KDE with the Gaussian kernel is given as
\begin{align}
\widehat{p}(\boldx) = \frac{1}{n(2\pi \sigma^2)^{d/2}} \sum_{k = 1}^n K(\boldx, \boldx_k),
\end{align}
where $K(\boldx, \boldx') = \exp \left( -\frac{\|\boldx - \boldx'\|^2}{2\sigma^2}\right)$.

KDE is easy to implement and scales well to the number of training samples. However, it is known that KDE suffers from the \emph{curse of dimensionality}, and thus, it may perform poorly when the dimensionality of samples are large. It has been empirically reported that the performance of KDE based outlier detection degrades for high-dimensional data \cite{KAIS:Hido+etal:2011}. Moreover, it is hard to select features in KDE.

\vspace{.1in}
\noindent {\bf One-class Support Vector Machine (OSVM):} \cite{book:Schoelkopf+Smola:2002}
One-class Support Vector Machine (SVM) is another widely used unsupervised outlier detection method. In this approach, OSVM regards samples located in the high density region as inliers. OSVM optimizes
\begin{align}
\min_{\boldalpha \in \mathbbR^{n}} & \hspace{0.3cm} \frac{1}{2} \sum_{k,k'=1}^n \alpha_k \alpha_{k'} K(\boldx_k, \boldx_{k'}) \nonumber \\
\text{s.t.} &\hspace{0.3cm} \boldalpha^{\top} \boldone = 1, 0 \leq \alpha_1, \ldots, \alpha_n \leq \frac{1}{n\nu},
\end{align}
where $\nu$ $(0 \leq \nu \leq 1)$ is a tuning parameter and $\boldone \in \mathbbR^n$ is the vector of all ones.

OSVM is a nonlinear kernel method, and hence the results are difficult to interpret in general. Moreover, since OSVM is a unsupervised outlier detection algorithm, the detection power can be lower if its assumptions are violated. 

\vspace{.1in}
\noindent {\bf Local Outlier Factor (LOF):} \cite{breunig2000lof}
LOF regards a sample as outlier if locally estimated density at the sample is low. The LOF score is defined as
\begin{align}
\text{LOF}_k(\boldx) = \frac{1}{K} \sum_{k = 1}^K \frac{\text{g}_k(\text{nearest}_k(\boldx))}{\text{g}_k(\boldx)},
\end{align}
where $\text{nearest}_k(\boldx)$ is the $k$-th nearest neighbor of $\boldx$ and $\text{g}_k(\boldx)$  is the inverse of the average distance from the $k$ nearest neighbors of $\boldx$.

It has been experimentally shown that LOF compares favorably with other unsupervised outlier detection algorithms. However, as in other outlier detection algorithms, it is not straightforward to interpret the decision. 

\subsection{Inlier-Based Outlier Detaction}
Here, we introduce three inlier-based outlier detection methods. It has been reported that the detection accuracy of inlier-based approach tends to be higher than that of unsupervised algorithms \cite{KAIS:Hido+etal:2011}.

\vspace{.1in}
\noindent{\bf $\ell_1$-Logistic Regression} \cite{bickel2007discriminative}
Let us define the posterior probability densities as
\begin{align}
\label{eq:logisticmodel}
\begin{split}
p(y = +1 | \boldx; \boldw)  &= \frac{1}{1 + \exp(-\boldw^\top \boldx)}, \\
p(y = -1 | \boldx; \boldw)  &= \frac{\exp(-\boldw^\top \boldx)}{1 + \exp(-\boldw^\top \boldx)},
\end{split}
\end{align}
where $\boldw \in \mathbbR^{d}$ contains the regression coefficients. 

Then, we can estimate the ratio function for $\boldx$ as
\begin{align*}
r(\boldx; \boldw) %&= \frac{p(y= -1)}{p(y = +1)} \frac{p(y = +1|\boldx)}{p(y = -1|\boldx)} \\
&=\frac{\ntr}{\nte}  \frac{p(y = +1|\boldx; \boldw)}{p(y = -1|\boldx; \boldw)} \\ 
                                           &=\frac{\ntr}{\nte} \exp(\boldw^\top \boldx).
\end{align*}
The model parameter is obtained by solving the following logistic regression problem:
\begin{align}
\label{eq:lr}
\min_{\boldw} &\hspace{0.1cm} \sum_{i = 1}^{\ntr + \nte} \log (1 + \exp(-y_i\boldw^\top \widetilde{\boldxtr}_i)) + \lambda\|\boldw\|_1,%\lambda_1 \sum_{i,j = 1}^{n} r_{ij} \|\boldw_i - \boldw_j\|_{2} \nonumber \\
%\phantom{\min_{\boldW} \hspace{0.1cm}  J(\boldW)} &\phantom{=}+ \lambda_2 \sum_{i = 1}^{\ntr}\|\boldw_i\|_1^2, %\nonumber \\
%\end{split}
\end{align}
where $\lambda > 0$ is a regularization parameter and $\|\boldw\|_1 = \sum_{i = 1}^d |w_i|$ is the $\ell_1$-regularizer. For this approach, we can select features through $\boldw$.  However, LR is a linear method, and thus, the outlier detection power can be lower if the data is nonlinearly behaved. 

\vspace{.1in}
\noindent{\bf Kullback-Leibler Importance Estimation Procedure (KLIEP)} \cite{sugiyama2008direct} 
KLIEP is a method to directly estimate the importance $r(\boldx) = \frac{\ptr(\boldx)}{\pte(\boldx)}$. Let us model the importance $r(\boldx)$ by the following model:
\begin{align}
r_{\boldalpha}(\boldx) = \sum_{\ell = 1}^{\nte} \alpha_\ell \exp \left( -\frac{\|\boldx - \boldxte_\ell\|^2}{2\tau^2}\right),
\end{align}
where $\boldalpha = [\alpha_1, \alpha_2, \ldots, \alpha_{\nte}]^\top$ is the model parameter and $\tau$ is a tuning parameter. Then, the optimization problem of KLIEP is given as
\begin{align*}
\max_{\boldalpha \in \mathbbR^b} & \hspace{0.3cm} \left[ \sum_{j = 1}^{\ntr} \log \left( r_{\boldalpha}(\boldxtr_j) \right) \right] \\
\text{s.t.} & \hspace{0.3cm} \sum_{\ell = 1}^{\nte} r_{\boldalpha}(\boldxte_i) = \nte, \alpha_1, \alpha_2, \ldots, \alpha_{\nte} \geq 0.
\end{align*}
This optimization problem is convex, and the global solution can be easily obtained by iterating gradient ascent and feasibility satisfaction alternatingly \cite{sugiyama2008direct}. KLIEP is convex and can deal with nonlinearity; however, it is not straightforward to select features in the KLIEP framework.

To select a set of features for KLIEP, Local kernel density-ratio (LoKDR) estimation \cite{azmandian2012local} could be applied in principle. However, LoKDR would not be able to estimate the density-ratio function $\frac{p(\boldx)}{p'(\boldx)}$ and, being a greedy approach, could suffer from low accuracy.

\vspace{.1in}
\noindent{\bf uLSIF/RuLSIF}  \cite{JMLR:Kanamori+etal:2009,yamada2011relative}
Let us first define the \emph{relative importance weight}~\cite{yamada2011relative}:
\begin{align}
\label{eq:relative-ratio}
w_{\alpha}(\boldx) = \frac{\pte(\boldx)}{(1-\beta)\pte(\boldx) + \beta \ptr(\boldx)},~0\leq \beta \leq 1,
\end{align}
where $\beta$ is a tuning parameter for controlling the \emph{adaptiveness} to the test distribution.

Let us model the relative importance weight by Eq \eqref{eq:linmodel}. Then, the parameters $\boldalpha$ in the model $\ratiomodel(\boldx;\boldalpha)$ are determined so that
the following expected squared-error $J$ is minimized:
\begin{align*}
  J(\boldtheta)
  \!&=\!  \frac{1}{2}\mathbbE_{q_{\beta}(\boldx)}
  \left[\left(\ratiomodel(\boldx;\boldtheta)-\relratio(\boldx)\right)^2\right]\\
  \!&=\!
  \frac{(1\!-\!\beta)}{2}\mathbbE_{\pte(\boldx)}\!\!\left[\ratiomodel(\boldx;\boldtheta)^2\right]
  \!\!+\!\frac{\beta}{2}\mathbbE_{\ptr(\boldx)}\!\!\left[\ratiomodel(\boldx;\boldtheta)^2\right] \\
  \!&\phantom{=}\! \!-\!\mathbbE_{\pte(\boldx)}\!\!\left[\ratiomodel(\boldx;\boldtheta)\right]
  \!\!+\!\mathrm{Const.},
\end{align*}
where $q_{\alpha}(\boldx) = (1-\alpha)\pte(\boldx) + \alpha\ptr(\boldx)$, and we used $\relratio(\boldx)q_{\alpha}(\boldx)=\pte(\boldx)$ in the third term (see supplemental materials for derivation).
% \leon{I think this equation is not quite right. We need to check for consistency of $\alpha$ as a sub-script for $w$.}

Approximating the expectations by empirical averages,
we obtain the following optimization problem:
\begin{align}
  \boldthetah=\argmin_{\boldalpha\in\mathbbR^{\nte}}
  \left[\frac{1}{2}\boldalpha^\top\boldHh\boldalpha-\boldhh^\top\boldalpha 
  +\frac{\nu}{2}\boldalpha^\top\boldalpha\right],
  \label{uLSIF-optimization-empirical}
\end{align}
where $\nu\boldalpha^\top\boldalpha/2$ is included to avoid overfitting,
and $\nu$ $(\ge0)$ denotes the regularization parameter.
$\boldHh$ is the $\nte\times\nte$ matrix with the $(\ell,\ell')$-th element
\begin{align*}
  \Hh_{\ell,\ell'} &= \frac{(1-\alpha)}{\nte}\sum_{i=1}^{\nte} \kappa( \boldxte_i, \boldxte_\ell)
                                                                                                    \kappa( \boldxte_i, \boldxte_{\ell'}) \\
                            &\phantom{=}+ \frac{\alpha}{\ntr} \sum_{j=1}^{\ntr} \kappa(\boldxtr_j, \boldxte_\ell)
                            	                                                                 \kappa(\boldxtr_j, \boldxte_{\ell'}).
\end{align*} 
% \begin{align*}
%   \Hh_{\ell,\ell'}&=
%   \frac{(1-\alpha)}{\nte}\sum_{i=1}^{\nte} \exp\left(-\frac{\unorms{\boldxte_i-\boldxte_\ell}}{2\tau^2}\right)  \exp \left(-\frac{\unorms{\boldxte_i-\boldxte_{\ell'}}}{2\tau^2}\right) \nonumber \\%K(\boldxte_i,\boldxte_\ell)K(\boldxte_i,\boldxte_{\ell'}) \nonumber \\
%   &\phantom{=} + \frac{\alpha}{\ntr} \sum_{j=1}^{\ntr} \exp\left(-\frac{\unorms{\boldxtr_j-\boldxte_\ell}}{2\tau^2}\right)  \exp \left(-\frac{\unorms{\boldxtr_j-\boldxte_{\ell'}}}{2\tau^2}\right).
%  K(\boldxtr_j,\boldxte_\ell)K(\boldxtr_j,\boldxte_{\ell'}).
 %\label{Hh}
% \end{align*}
$\boldhh$ is the $\nte$-dimensional vector with the $\ell$-th element
$  \hh_{\ell}=\frac{1}{\nte}\sum_{i=1}^{\nte} \kappa(\boldxte_i, \boldxte_\ell)$.  
% \begin{align*}
%   \hh_{\ell}=\frac{1}{\nte}\sum_{i=1}^{\nte} \exp\left(-\frac{\unorms{\boldxte_i-\boldxte_\ell}}{2\tau^2}\right).  %K(\boldxte_i,\boldxte_\ell).
%   \label{hh}
% \end{align*}
%It is easy to confirm that the solution of Eq.\eqref{uLSIF-optimization-empirical}
%can be \emph{analytically} obtained as
Then the solution to Eq.~\eqref{uLSIF-optimization-empirical} can be \emph{analytically} obtained as
\begin{align}
  \widehat{\boldalpha}=(\boldHh+\nu\boldI)^{-1}\boldhh,
\label{boldalphah-uLSIF}
\end{align}
where $\boldI$ is the $\nte \times \nte$-dimensional identity matrix.

Similar to KLIEP, since uLSIF uses entire features, it may perform poorly for high-dimensional data if there exist many noisy features. Moreover, it is hard to select a set of features of the outliers.

\begin{figure*}[t!]
\begin{center}
\begin{minipage}[t]{0.235\linewidth}
\centering
  {\includegraphics[width=0.99\textwidth]{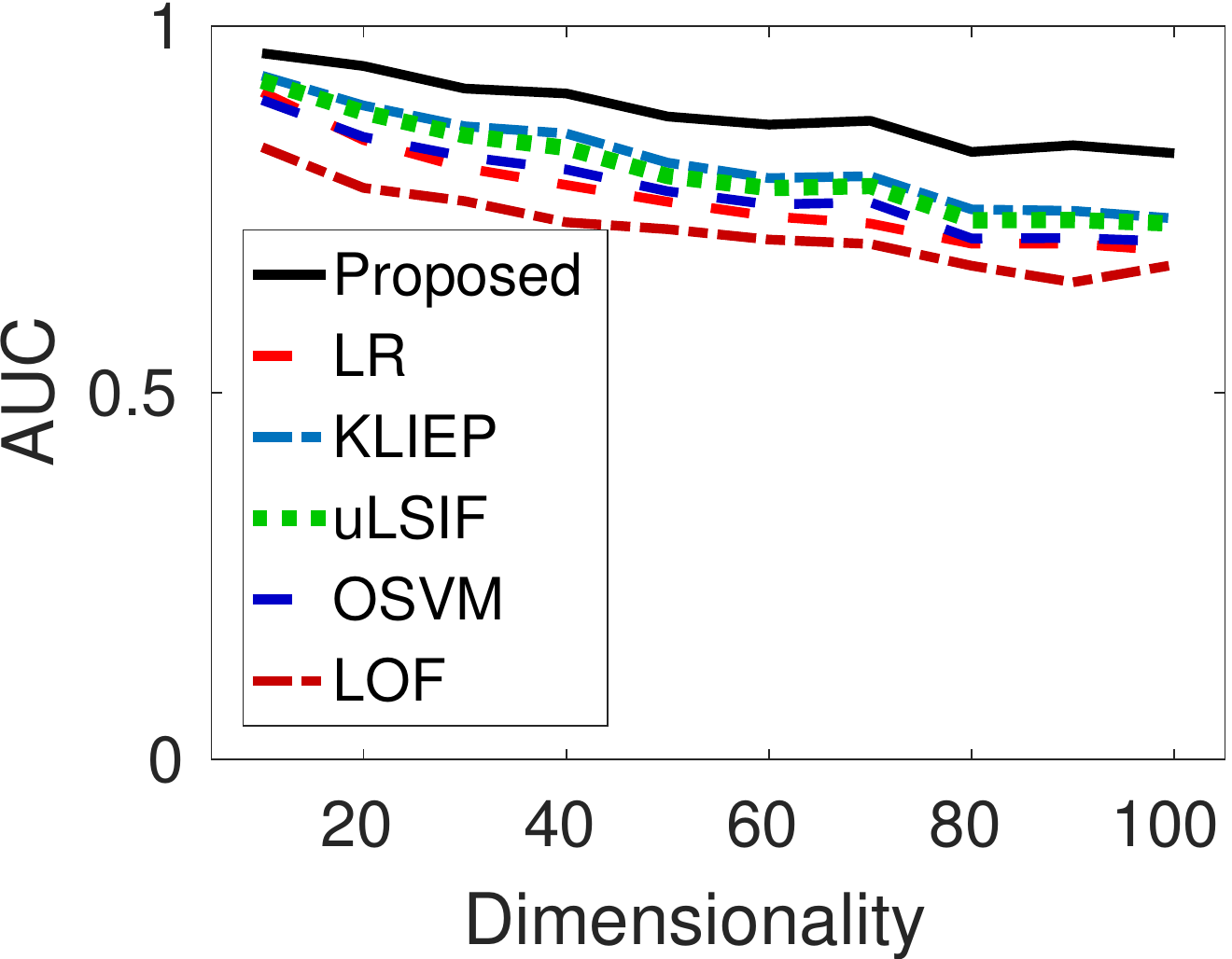}} \\ \vspace{-0.10cm}
(a) AUC.
\end{minipage}
\begin{minipage}[t]{0.255\linewidth}
\centering
  {\includegraphics[width=0.99\textwidth]{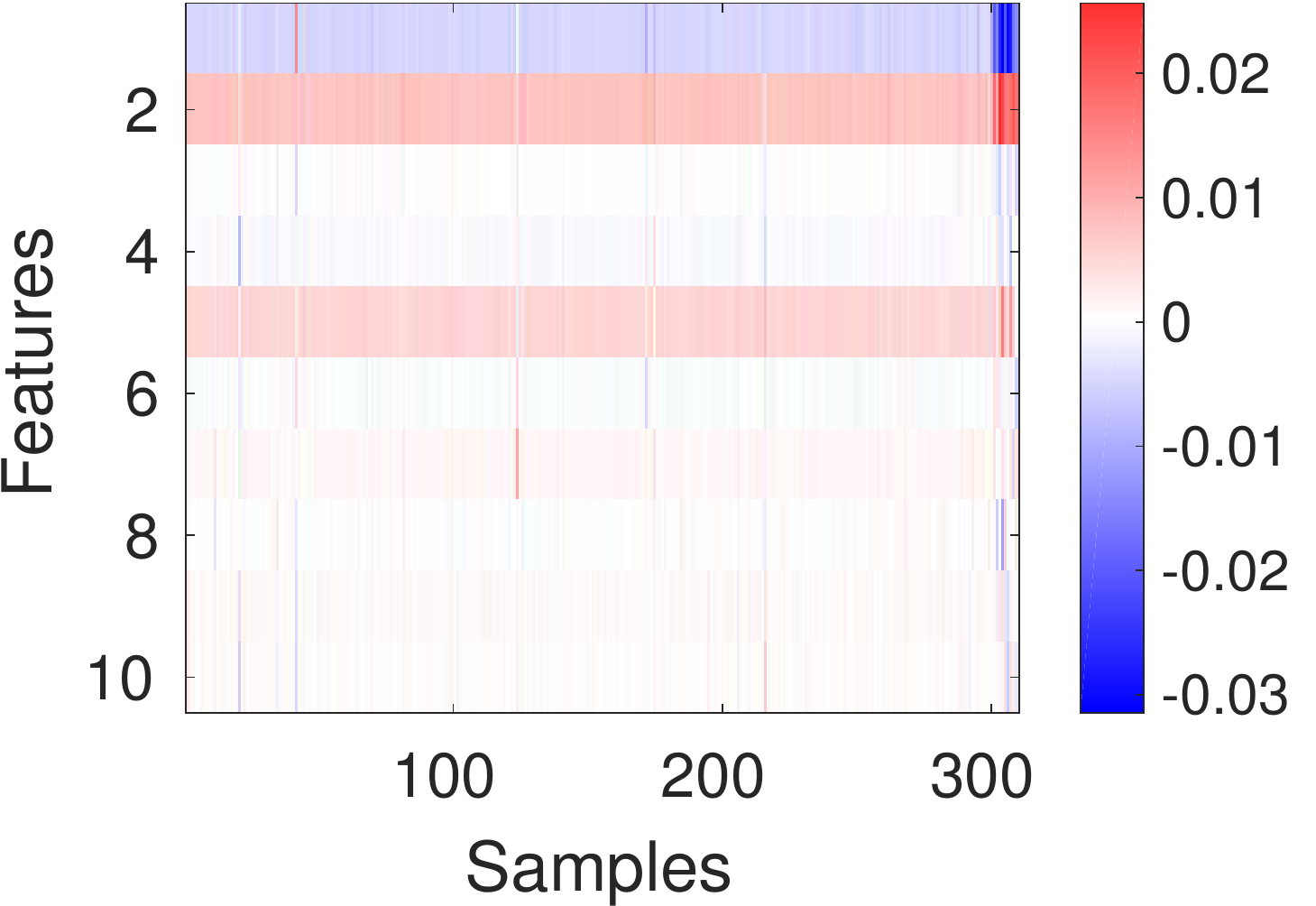}} \\ \vspace{-0.10cm}
(b) The learned coefficient matrix.
\end{minipage}
\begin{minipage}[t]{0.235\linewidth}
\centering
  {\includegraphics[width=0.99\textwidth]{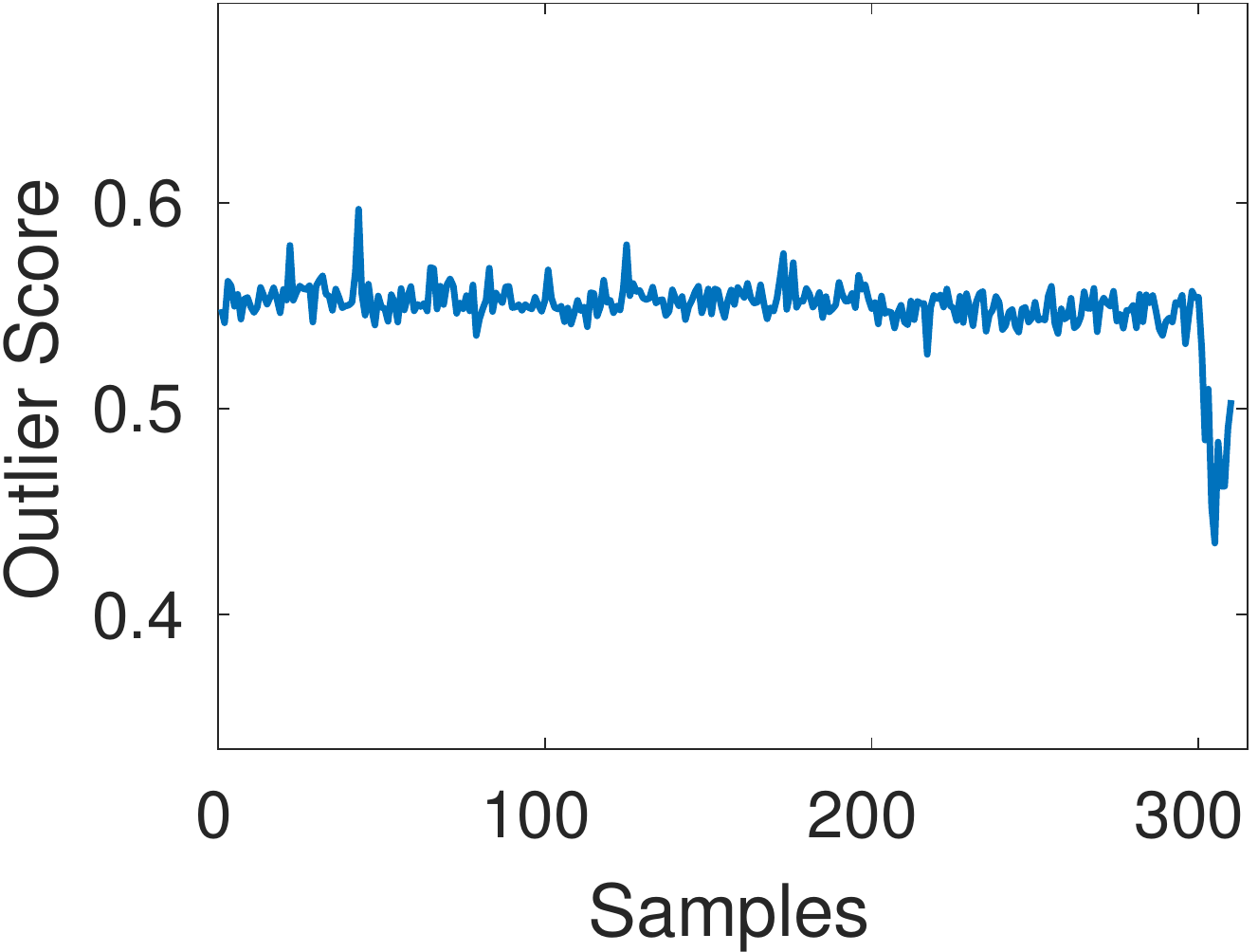}} \\ \vspace{-0.10cm}
(c) The learned outlier score.
\end{minipage}
\begin{minipage}[t]{0.245\linewidth}
\centering
  {\includegraphics[width=0.99\textwidth]{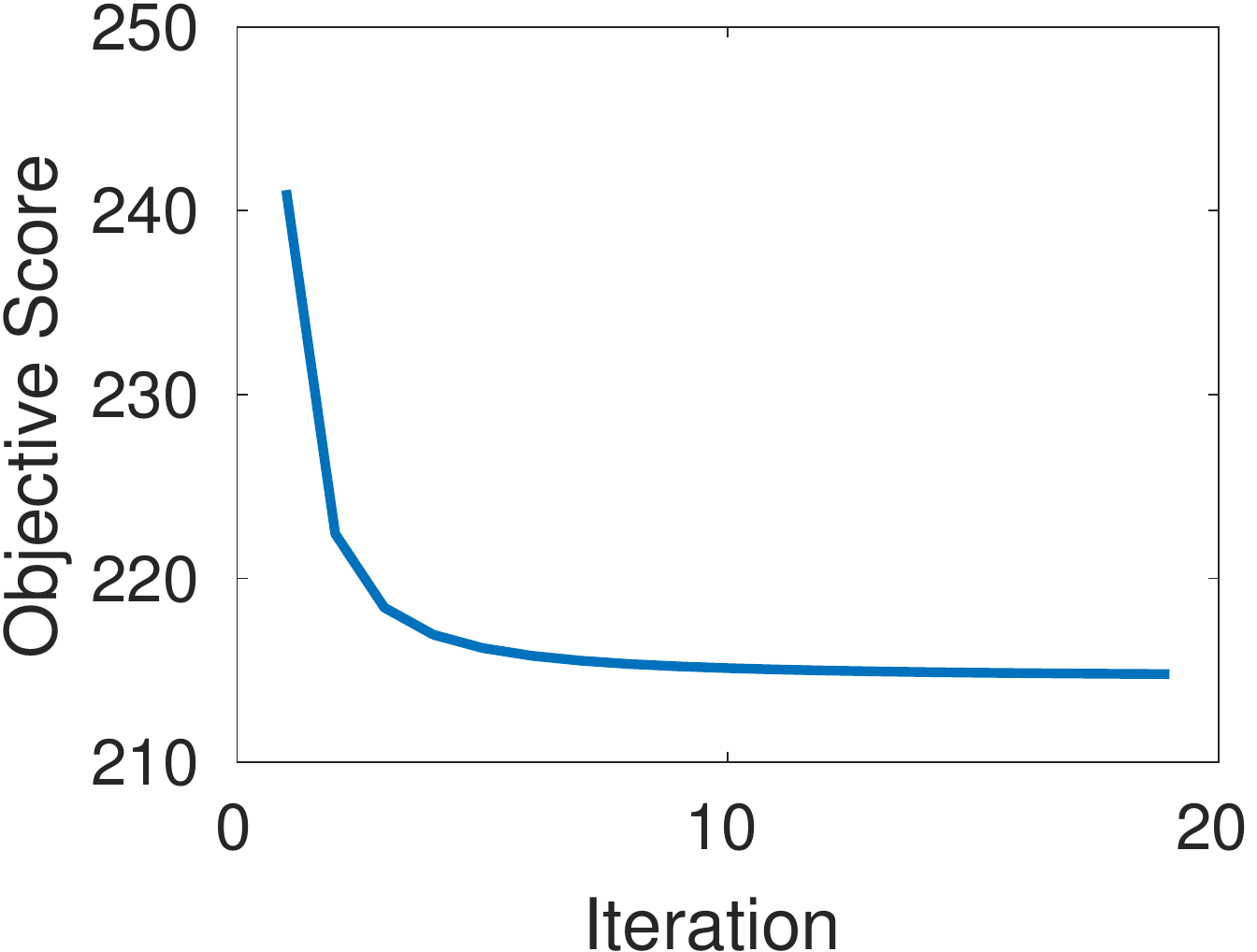}} \\ \vspace{-0.10cm}
(d) Convergence.
\end{minipage}
\centering
\caption{Synthetic data results. (a) AUC. (b). The learned coefficient matrix for the synthetic data, where $\lambda_1 = 0.1$ and $\lambda_2 = 1$.  (c) The learned outlier score. (d) Convergence of the proposed method; objective score as a function of number of iterations.}
\label{fig:synthetic}
\end{center}
\end{figure*}

%\begin{figure}[t!]
%\begin{center}
%\begin{minipage}[t]{0.9\linewidth}
%\centering
%  {\includegraphics[width=0.99\textwidth]{fval.eps}} \\ \vspace{-0.10cm}
%\end{minipage}
%\centering
%\caption{Convergence of the proposed method; objective score as a function of number of iterations.}
%\label{fig:fval}
%\end{center}
%\vspace{-.2in}
%\end{figure}

\begin{table*}[t!]
\centering
\caption{Experimental results of outlier detection
  for benchmark datasets. The figures are the mean AUC scores (with standard deviations in the brackets) over
  $100$ trials. The best method having the highest mean AUC score and
  comparable methods according to the \emph{t-test} at the significance
  level $5\%$ are boldfaced.
}
\label{tb:outlier_result}
\begin{tabular}{|@{\ }l@{\ }||l@{}c|l@{}r|l@{}r|l@{}r|l@{}r|l@{}r|l@{}r|l@{}r}
\hline
Datasets  
&\multicolumn{2}{c|}{
OSVM
} 
& \multicolumn{2}{c|}{
LOF
}
& \multicolumn{2}{c|}{
$\ell_1$-LR
} 
& \multicolumn{2}{c|}{
  KLIEP
} 
& \multicolumn{2}{c|}{
  uLSIF
} 
& \multicolumn{2}{c|}{
  Proposed
} \\
\hline
Breast Cancer & 0.632 & ~(0.114) & 0.541  & ~(0.104) & 0.579 & ~(0.150) & 0.624 & ~(0.133) & 0.641 & ~(0.125)  & {\bf 0.739} & ~(0.117)  \\ \hline
Diabetes & 0.683 & ~(0.047) & 0.640 & ~(0.051) & 0.677 & ~(0.129) & 0.711 & ~(0.074) & 0.718 & ~(0.047)  & {\bf 0.761} & ~(0.096)  \\ \hline
German & 0.580 & ~(0.047) & 0.583 & ~(0.055) & 0.592 & ~(0.093) & 0.604 & ~(0.059) & 0.589 & ~(0.050)  & {\bf 0.692} & ~(0.045)  \\ \hline
Heart & 0.724 & ~(0.064) & 0.589 & ~(0.076) & 0.748 & ~(0.126) & 0.792 & ~(0.082) & {\bf 0.796} & ~(0.057)  & 0.743  & ~(0.073)  \\ \hline
Flare solar & 0.681 & ~(0.044) & 0.415 & ~(0.035) & 0.672 & ~(0.059) & 0.668 & ~(0.063) & 0.693 & ~(0.045)  & {\bf 0.781} & ~(0.044)  \\ \hline
Vehicle  & 0.590 & ~(0.048) & 0.552 & ~(0.069) & 0.563 & ~(0.119) & 0.562 & ~(0.127) & 0.594 & ~(0.096)  & {\bf 0.770} & ~(0.049)  \\ \hline
%Wine  &  & ~() &  & ~() &  & ~() &  & ~() &  & ~()  &  & ~()  \\ \hline
Glass  & 0.394 & ~(0.088) & 0.465 & ~(0.092) & {\bf 0.520} & ~(0.153) & 0.455 & ~(0.144) & 0.389 & ~(0.092)  & 0.468  & ~(0.104)  \\ \hline \hline
Average  & 0.612 & ~(0.123) & 0.565 & ~(0.088) & 0.622 & ~(0.142) & 0.631 & ~(0.143) & 0.631 & ~(0.143)  & {\bf 0.712} & ~(0.128)  \\ \hline 
\end{tabular}
\end{table*}

\section{Experiments}
In this section, we experimentally evaluate our proposed method by using synthetic and  benchmark datasets. We consider the task of finding anomalies in a test dataset based on a training dataset which only includes inliers. 

We compared our proposed method with unsupervised outlier detection methods (OSVM, LOF) and inlier-based outlier detection methods ($\ell_1$-LR, KLIEP, uLSIF). For OSVM, we used the libSVM implementation \cite{CC01a} with the Gaussian kernel, where the Gaussian width is set to the median distance between samples. For LOF, KLIEP, and uLSIF, we used publicly available codes \footnote{\url{https://bitbucket.org/gokererdogan/outlier-detection-toolbox/}}\footnote{\url{http://www.ms.k.u-tokyo.ac.jp/software.html}}. For $\ell_1$-LR, we used the Matlab function \texttt{lassoglm}. For the proposed algorithm, we empirically set $\lambda_1 = 0.1$ and $\lambda_2 = 1$ for all experiments.

To compare fairly all the unsupervised and inlier-based outlier detection methods, we use the inliers in addition to test samples for OSVM and LOF. More specifically, in OSVM, we first pooled all the inliers and test samples, and ran OSVM on the resulting new test dataset. For LOF, we used the inliers to build the model. With these modifications, performance of OSVM and LOF improved significantly.

Throughout these experiments, when evaluating the performance of outlier detection methods, we need to evaluate both the \emph{detection rate} (i.e., the amount of true anomalies an outlier detection algorithm can find) and the \emph{detection accuracy} (i.e., the amount of true inliers an outlier detection algorithm misjudges as anomalies). Since there is a trade-off between the detection rate and the detection accuracy, we use the \emph{area under the ROC curve} (AUC) \cite{bradley1997use} as our accuracy metric.

\subsection{Synthetic Data}
%\subsubsection{Synthetic Datasets}
\label{sec:experiment-outlier-artificial}
First, we illustrate how the proposed method behaves in outlier detection scenarios using synthetic datasets. 

Let 
\begin{align*}
  \Pnu&=N(0,\boldI_\inputdim),\\
  \Pde & =\left\{ \begin{array}{ll}
N(\boldzero_d,\boldI_d) & (\text{Inliers}) \\
N(\boldmu,\boldI_{d})&  (\text{Outliers})\\
\end{array} \right.,
\end{align*}
where $\boldmu = [3, -2, 0, \ldots, 0]^\top \in \mathbbR^{d}$.

Here, the samples drawn from $P'$ are regarded as inliers, while the samples drawn from $P$ are regarded as the test set. For the inlier dataset, we used $\nte=200$  samples. For the test dataset, we used $100$ inliers and $10$ outliers (total $\ntr = 110$ samples). We run the outlier detection experiments 100 times by randomly selecting inliers and outliers and report the average AUC scores.

Figure \ref{fig:synthetic}(a) shows the comparison between the proposed and existing methods. In the figure, we evaluated algorithms with different numbers of features (i.e., $d = 10, 20, \ldots, 100$).  As can be observed, when the dimensionality is small, all methods can detect outliers.  However, for high-dimensional case, the proposed method highly outperforms existing approaches. Note that even though the $\ell_1$-LR method can select features, as a linear method it tends to perform poorly. Figure \ref{fig:synthetic}(b) shows a learned coefficient matrix. The coefficients corresponding to outliers (sample indices 301 to 310) tend to have higher absolute values. In addition, \ref{fig:synthetic}(c) shows the learned outlier score Eq.\eqref{eq:outlier_score}, which clearly shows that the outlier scores of outliers have lower values. Finally, Figure \ref{fig:synthetic}(d) shows the convergence of the proposed method. The objective score converges within 20 iterations. 

\subsection{Benchmark Data}
Next, we evaluated the proposed method with a number of UCI datasets \cite{Lichman:2013}. In two-class datasets, we regard samples in the positive class as inliers and samples in negative classes as outliers. For multi-class datasets, we regard samples in class "1" as outliers and the samples of rest of the classes as inliers. We randomly took half of the samples from the inlier class and used them as model samples. Then we randomly used the rest of the samples from the same inlier class and 10 samples from outliers as a test dataset.

The mean and standard deviation of the AUC scores over 100 runs are summarized in Table\ref{tb:outlier_result}. As can be observed, the proposed algorithm compares favorably with the earlier algorithms.

\section{Conclusion}
We proposed an inlier-based outlier detection approach for high-dimensional data. More specifically, we proposed a local density-ratio function to measure plausibility of being an outlier, where the model parameters are efficiently learned by convex optimization. Since the proposed method learns a (locally) linear model, we can impose regularization to perform feature selection for the outliers. In both  synthetic and benchmark datasets, we showed that the proposed approach outperforms existing state-of-the-art algorithms.

\bibliographystyle{unsrt}
\bibliography{main}

\end{document}